\documentclass{article}
\usepackage[numbers,sort&compress]{natbib}
\usepackage{spconf,amsmath,graphicx,hyperref}
\usepackage[utf8]{inputenc} 
\usepackage[T1]{fontenc}    
\usepackage{url}            
\usepackage{booktabs}       
\usepackage{amsfonts}       
\usepackage{nicefrac}       
\usepackage{microtype}      
\usepackage{graphicx}
\usepackage{colortbl}
\usepackage{array}
\usepackage{caption}
\usepackage{makecell}
\usepackage{wrapfig}
\usepackage{subcaption} 
\usepackage{graphicx}
\usepackage[numbers]{natbib}  
\usepackage[dvipsnames]{xcolor}

\usepackage{pifont} 

\usepackage{wrapfig}
\usepackage{multirow}
\usepackage{multicol}
\usepackage{listings} 
\usepackage{titletoc} 
\usepackage{fancyvrb}
\usepackage{tcolorbox} 

\definecolor{uptri}{RGB}{38,114,168}
\definecolor{dntri}{RGB}{178,34,52}

\definecolor{lightgray}{rgb}{0.83, 0.83, 0.83}
\definecolor{Gray}{gray}{0.6}
\definecolor{aliceblue}{rgb}{0.94, 0.97, 1.0}
\definecolor{mistyrose}{rgb}{1.0, 0.89, 0.88}
\definecolor{backcolour}{rgb}{0.95,0.95,0.92}

\definecolor{gainblue}{RGB}{40,100,190}
\definecolor{lossred}{RGB}{190,60,60}
\definecolor{neutralgray}{RGB}{110,110,110}
\definecolor{modelhighlight}{RGB}{235,240,250}  
\usepackage{xcolor}
\definecolor{leftcolor}{HTML}{74B9D3}
\definecolor{rightcolor}{HTML}{9A9A9A}

\newcommand{\newpara}[1]{\vspace{2pt}\noindent\textbf{#1}}

\usepackage[accsupp]{axessibility}
\usepackage{hyperref}
\usepackage{cleveref}

\crefname{equation}{Eq.}{Eqs.}
\Crefname{equation}{Equation}{Equations}

\crefname{figure}{Fig.}{Figs.}
\Crefname{figure}{Figure}{Figures}

\crefname{table}{Tab.}{Tabs.}
\Crefname{table}{Table}{Tables}

\crefname{section}{Sec.}{Secs.}
\Crefname{section}{Section}{Sections}

\crefname{algorithm}{Alg.}{Algs.}
\Crefname{algorithm}{Algorithm}{Algorithms}

\title{Uncovering Ordinal-Matching Bias in Audio-Visual LLMs}
\name{Jihoo Jung$^{1}$, Youngjoon Jang$^{2}$, Hyebin Cho$^{1}$, Suho Yoo$^{1}$, Joon Son Chung$^{1}$}
\address{$^{1}$Korea Advanced Institute of Science and Technology, South Korea \\ $^{2}$University of Oxford, United Kingdom}

\begin{document}
%
\maketitle
\begin{abstract}
This work aims to improve how audio-visual large language models (AVLLMs) associate speech with the correct visible speaker in multi-speaker scenes. We find that current AVLLMs frequently fail at this task, and analyze the nature of these failures.
To this end, we construct a synthetic diagnostic dataset in which multiple visible speakers each utter a single word. Analysis on this corpus reveals a consistent error pattern across three recent open-source AVLLMs: models attribute utterances by simply matching the order of spoken sentences with the left-to-right, top-to-bottom arrangement of visible faces, rather than relying on audio-visual cues such as lip synchronization. We term this behavior \emph{ordinal-matching bias}. We further show that this bias can be substantially mitigated through a simple remedy, Ordinal-Decoupled Fine-Tuning (OD-FT), in which models are fine-tuned on synthetic videos where spatial positions of speakers and speaking order are independently randomized. Despite using only 400 synthetic training videos, OD-FT not only suppresses ordinal-matching bias but also improves audio-visual understanding on real-world videos, yielding average gains of 8.27\% for Qwen2.5-Omni and 2.57\% for video-SALMONN2+ across three audio-visual benchmarks.

\end{abstract}
\section{Introduction}
\label{sec:intro}

Audio-visual large language models (AVLLMs)~\cite{tang2025video,cui2026minicpm,xu2025qwen25omnitechnicalreport,xu2025qwen3omnitechnicalreport} extend large language models to jointly perceive vision and audio, enabling complex reasoning over sounding videos. These advances have also spurred interest in understanding the internal mechanisms of AVLLMs~\cite{jung2026probing, yoo2026nature}. As conversation-rich videos, including films, and daily vlogs, account for a significant fraction of contemporary video data, speaker-utterance attribution--attributing each utterance to the correct visible speaker--has become an essential capability. Despite strong performance on general audio-visual benchmarks, current AVLLMs perform poorly on speaker-utterance attribution, frequently assigning an utterance to the wrong visible person~\cite{jung2026says, tang2026d,chen2026diadem,nguyen2025see}.

Evidence from closely related architectures suggests that such systematic failures often stem from underlying biases rather than random errors. LLMs, for instance, become unreliable when reasoning over long contexts because their attention is biased toward certain positions regardless of content~\cite{liu2024lost,tang2024found,an2024make}. Similarly, large vision-language models (LVLMs) are unreliable over multiple images or long videos largely because their predictions depend heavily on where the visual inputs are positioned rather than on their actual content~\cite{tian2025identifying,xia2026videolevelgauge}. This raises the possibility that speaker-attribution errors in AVLLMs follow a similar pattern: when associating auditory signals with visible speakers, the model may rely on spurious cues--the spatial arrangement of faces and the temporal order of utterances--instead of genuine audio-visual correspondence.

\begin{figure}[t]
\centering
\includegraphics[width=\linewidth]{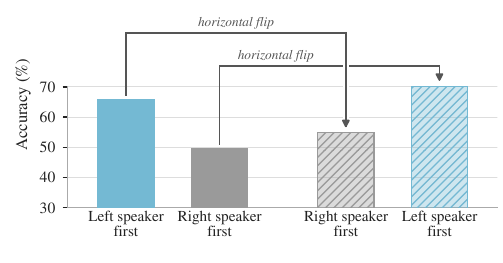}
\vspace{-10mm}
\caption{\textbf{Position bias on curated two-speaker clips with video-SALMONN2+.} Accuracy on ``Who spoke first?'' is substantially higher when the first speaker is on the left. This gap persists even for the same videos after horizontal flipping, indicating a strong bias.}
\vspace{-7mm}
\label{fig:init}
\end{figure}

To validate this hypothesis, we conduct a preliminary experiment on two-person conversation clips, where the two speakers appear on the left and right halves of the frame, using the dataset of~\cite{xie2026socialomni}.
Each clip is evaluated under two conditions: the original version and a horizontally flipped version, which reverses the visual spatial position of the speakers in each frame while preserving all semantic content. In both cases, we prompt the model with ``Who spoke first?'' and restrict the answer choices to ``left'' or ``right.''. \cref{fig:init} reveals a systematic position-related bias. First, in both the original videos (solid bars) and the flipped videos (hatched bars), accuracy is substantially higher for left-speaker-first clips \textcolor{leftcolor}{(blue)} than for right-first clips \textcolor{rightcolor}{(gray)}. Second, when comparing the original and horizontally flipped versions of the same video, the model's accuracy changes substantially: it consistently achieves higher accuracy whenever the first speaker happens to be positioned on the left. Together, these findings reveal a \emph{left bias}: the model inherently attributes the initial speech to the left-positioned speaker, motivating further analysis.

To analyze such behavior more systematically, we construct a controlled, synthetic video corpus featuring several speakers, each uttering a single word. By evaluating three recent open-source AVLLMs~\cite{xu2025qwen25omnitechnicalreport,tang2025video,xu2025qwen3omnitechnicalreport}, we observe that all models consistently fail to correctly associate the visible speakers with their corresponding spoken utterances. Crucially, these failures are not random: they follow a systematic pattern, which we term \textbf{ordinal-matching bias}. Rather than relying on genuine audio-visual cues such as lip synchronization, the models tend to link the $k$-th temporal utterance with the $k$-th speaker in the visual positional order. For example, as depicted in \cref{fig:main}, when queried about the panda located in the top-left position (the first visual subject), the model routinely associates it with the first spoken utterance (``Egypt''), even though the panda actually produces the second utterance (``Mexico'').

This diagnosis suggests a simple remedy: if AVLLMs rely on the spurious correspondence between visual position and utterance order, then training examples that explicitly decouple these two orders should discourage such a shortcut. Based on this intuition, we introduce \textbf{Ordinal-Decoupled Fine-Tuning (OD-FT)}. In OD-FT, we fine-tune two AVLLMs~\cite{xu2025qwen25omnitechnicalreport,tang2025video} on a targeted set of just 400 synthetic videos where the speaker spatial positions and the temporal utterance orders are fully randomized. OD-FT substantially raises speaker-utterance attribution accuracy, effectively eliminating the ordinal-matching bias. Importantly, the gains are not confined to our synthetic setting: on three real-world audio-visual benchmarks, OD-FT improves Qwen2.5-Omni~\cite{xu2025qwen25omnitechnicalreport} by 8.27\% and video-SALMONN2+~\cite{tang2025video} by 2.57\% on average.

\section{Identifying Ordinal-Matching Bias}
\label{sec:identify}

We construct a synthetic diagnostic corpus (\cref{sec:diag_setup}) to investigate failure modes in speaker--utterance attribution, identify a systematic failure pattern that we term ``ordinal-matching bias'' (\cref{sec:existence}), and provide causal evidence of such bias (\cref{sec:causal}).

\subsection{Analysis Setup}
\label{sec:diag_setup}

\begin{figure}[t]
\centering
\includegraphics[width=0.95\linewidth]{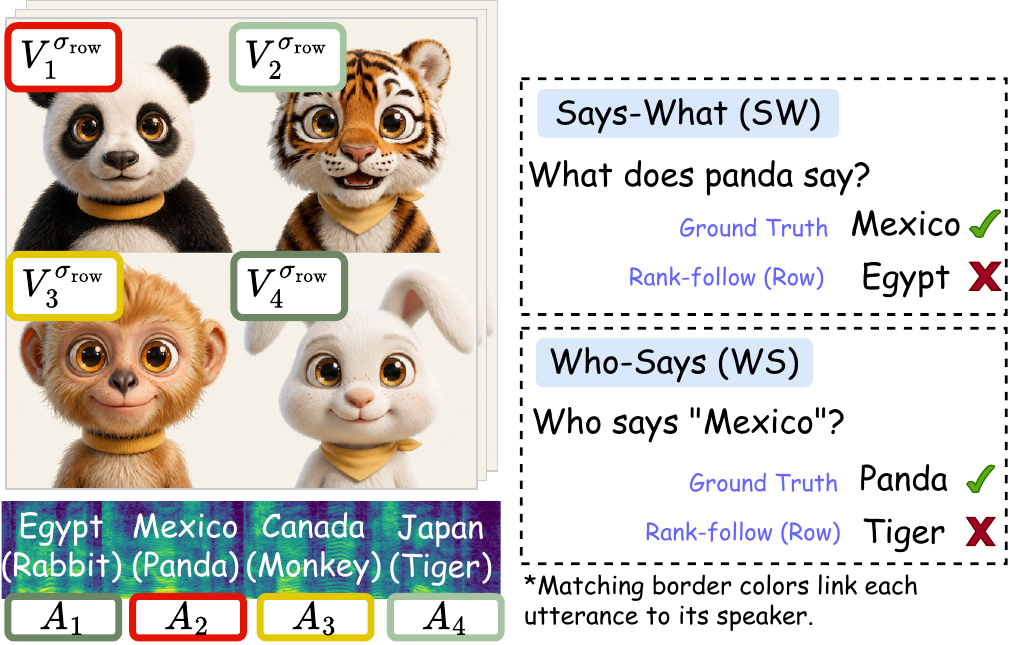}
\vspace{-3mm}
\caption{\textbf{Illustration of ordinal-matching bias.}
\textbf{SW:} The model predicts that the 1st on-screen speaker (``panda'') produced the 1st utterance (``Egypt''), rather than its true utterance (``Mexico'').
\textbf{WS:} The model predicts that the 2nd utterance (``Mexico'') was produced by the 2nd on-screen speaker (``tiger''), rather than its true source (``panda'').}

\label{fig:main}
\vspace{-5mm}
\end{figure}

\newcommand{\psize}[1]{\,\textcolor{neutralgray}{\scriptsize #1}}
\newcommand{\hl}{\cellcolor{modelhighlight}}
\begin{table*}[t]
\centering
\footnotesize
\setlength{\tabcolsep}{4pt}
\renewcommand{\arraystretch}{1.05}
\caption{\textbf{Existence of ordinal-matching bias.} For open-source AVLLMs, accuracy is low while $\mathrm{OM}_{\sigma_{\mathrm{row}}}$ is high, i.e., models attribute the $k$-th utterance to the speaker at the $k$-th position in row-major order, confirming ordinal-matching bias.}
\vspace{-3mm}
\begin{tabular}{@{}ll ccc ccc ccc@{}}
\toprule
\multirow{2}{*}{\textbf{Model}} & \multirow{2}{*}{\textbf{Metric}}
& \multicolumn{3}{c}{$N{=}4$} & \multicolumn{3}{c}{$N{=}5$} & \multicolumn{3}{c}{$N{=}6$} \\
\cmidrule(lr){3-5} \cmidrule(lr){6-8} \cmidrule(lr){9-11}
& & SW & WS & \textit{Chance} & SW & WS & \textit{Chance} & SW & WS & \textit{Chance} \\
\midrule
\multirow{3}{*}{Qwen2.5-Omni~\cite{xu2025qwen25omnitechnicalreport}\psize{7B}}
& Accuracy $\uparrow$& 26.9 & 25.9 & \textit{25.0} & 23.7 & 24.1 & \textit{20.0} & 18.9 & 19.3 & \textit{16.7} \\
\cmidrule(lr){2-11}
& \hl $\mathrm{OM}_{\sigma_{row}}$  & \hl \textbf{70.4} & \hl \textbf{75.0} & \hl \textit{25.0} & \hl \textbf{54.6} & \hl \textbf{62.2} & \hl \textit{20.0} & \hl \textbf{48.9} & \hl \textbf{53.7} & \hl \textit{16.7} \\
& $\mathrm{OM}_{\sigma_{col}}$ & 57.0 & 53.1 & \textit{25.0} & 43.6 & 45.4 & \textit{20.0} & 47.8 & 47.2 & \textit{16.7} \\
\cmidrule(l{0pt}r{0pt}){1-11}
\multirow{3}{*}{video-SALMONN2+~\cite{tang2025video}\psize{7B}}
& Accuracy $\uparrow$& 34.0 & 27.8 & \textit{25.0} & 27.7 & 25.0 & \textit{20.0} & 27.0 & 24.7 & \textit{16.7} \\
\cmidrule(lr){2-11}
& \hl $\mathrm{OM}_{\sigma_{row}}$ & \hl \textbf{67.4} & \hl 53.0 & \hl \textit{25.0} & \hl \textbf{60.6} & \hl \textbf{49.4} & \hl \textit{20.0} & \hl \textbf{49.5} & \hl \textbf{46.3} & \hl \textit{16.7} \\
& $\mathrm{OM}_{\sigma_{col}}$ & 51.5 & \textbf{54.9} & \textit{25.0} & 45.4 & 41.4 & \textit{20.0} & 40.6 & 40.6 & \textit{16.7} \\
\cmidrule(l{0pt}r{0pt}){1-11}
\multirow{3}{*}{Qwen3-Omni~\cite{xu2025qwen3omnitechnicalreport}\psize{30B-A3B}}
& Accuracy $\uparrow$& 25.9 & 28.2 & \textit{25.0} & 20.2 & 25.8 & \textit{20.0} & 18.7 & 19.4 & \textit{16.7} \\
\cmidrule(lr){2-11}
& \hl $\mathrm{OM}_{\sigma_{row}}$ & \hl \textbf{68.4} & \hl \textbf{59.4} & \hl \textit{25.0} & \hl \textbf{54.6} & \hl \textbf{53.7} & \hl \textit{20.0} & \hl \textbf{49.7} & \hl \textbf{47.3} & \hl \textit{16.7} \\
& $\mathrm{OM}_{\sigma_{col}}$ & 42.5 & 37.8 & \textit{25.0} & 43.5 & 33.2 & \textit{20.0} & 37.0 & 31.0 & \textit{16.7} \\

\midrule
\multirow{3}{*}{Gemini 3.7 Flash~\cite{gemini37flash}}
& Accuracy $\uparrow$ & 97.0 & 98.1 & \textit{25.0} & 97.0 & 96.8 & \textit{20.0} & 95.6 & 95.6 & \textit{16.7} \\
\cmidrule(lr){2-11}
& \hl $\mathrm{OM}_{\sigma_{row}}$ & \hl \textbf{24.6} & \hl \textbf{24.9} & \hl \textit{25.0} & \hl 19.8 & \hl 19.5 & \hl \textit{20.0} & \hl \textbf{16.7} & \hl 16.5 & \hl \textit{16.7} \\
& $\mathrm{OM}_{\sigma_{col}}$ & 24.5 & 24.6 & \textit{25.0} & 19.8 & \textbf{19.9} & \textit{20.0} & 16.1 & \textbf{16.6} & \textit{16.7} \\

\bottomrule
\end{tabular}
\vspace{-6mm}
\label{tab:cardinality}
\end{table*}

\newpara{Diagnostic corpus.}
As illustrated in \cref{fig:main}, we construct a diagnostic corpus of synthetic videos. Each video features $N \in \{4, 5, 6\}$ distinct animal speakers placed at fixed positions, with each speaker uttering a single country name in a randomized sequence.\footnote{This choice avoids potential confounds associated with human demographic attributes, such as gender.} The static speaker images are generated using Gemini3.1-Flash-Lite-Image~\cite{gemini31flashliteimage} and subsequently animated with synchronized speech utilizing a recent talking-face generation model~\cite{meng2026echomimicv3}. 
Formally, each video is represented as a set of $N$ speaker-utterance pairs mapping spatially localized speakers to their temporal speech sequence under a visual positional order $\sigma$: $\{(V_i^\sigma, A_{j})\}_{i,j=1}^N$. Here, $V_i^\sigma$ represents the visual speaker located at rank $i \in \{1, \dots, N\}$ under ordering $\sigma$. We employ two such positional orders: \emph{row-major} $\sigma_{\mathrm{row}}$ (e.g., top-left $\rightarrow$ top-right $\rightarrow$ bottom-left $\rightarrow$ bottom-right) and \emph{column-major} $\sigma_{\mathrm{col}}$ (e.g., top-left $\rightarrow$ bottom-left $\rightarrow$ top-right $\rightarrow$ bottom-right). Furthermore, $A_j$ denotes the $j$-th spoken utterance in the audio sequence $\mathbf{A} = (A_1, A_2, \dots, A_N)$. For example, the $N=4$ video in \cref{fig:main} under a row-major order is formulated as $\{(V_1^{\sigma_{\mathrm{row}}}, A_2), (V_2^{\sigma_{\mathrm{row}}}, A_4), (V_3^{\sigma_{\mathrm{row}}}, A_3), (V_4^{\sigma_{\mathrm{row}}}, A_1)\}$, indicating, for instance, that the top-left speaker speaks second and the bottom-right speaks first.

\newpara{Tasks.}
We evaluate two complementary open-ended QA tasks. In \emph{Says-What} (SW), the model is given a target speaker $V_i^\sigma$, identify the utterance $A_j$ spoken by $V_i^\sigma$. Conversely, in \emph{Who-Says} (WS), the model is given a target utterance $A_j$, identify the visual speaker $V_i^\sigma$ who uttered it.

\newpara{Models.}
We evaluate three recent open-source AVLLMs-Qwen2.5-Omni (7B) ~\cite{xu2025qwen25omnitechnicalreport}, video-SALMONN2+ (7B)~\cite{tang2025video}, and Qwen3-Omni (30B-A3B)~\cite{xu2025qwen3omnitechnicalreport}-and one proprietary model, Gemini 3.7 Flash~\cite{gemini37flash}.

\newpara{Metrics.}
We report two metrics. (i)~\emph{Accuracy} measures whether the model prediction matches the ground-truth speaker-utterance attribution. (ii)~\emph{Ordinal-matching rate} ($\mathrm{OM}_{\sigma}$) quantifies the model's tendency to associate the $k$-th speaker $V_k^\sigma$ with the $k$-th utterance $A_k$, regardless of the true correspondence. Formally, let $\mathcal{D}$ be a set of evaluation queries, $\hat{y}_q$ the model prediction for query $q \in \mathcal{D}$, and $\tilde{y}_q^\sigma$ the prediction dictated by the ordinal-matching rule ($V_k^\sigma \leftrightarrow A_k$). The ordinal-matching rate is defined as
\begin{equation*}
\mathrm{OM}_{\sigma} = \frac{1}{|\mathcal{D}|} \sum_{q \in \mathcal{D}} \mathbf{1}\left[ \hat{y}_q = \tilde{y}_q^{\sigma} \right].
\vspace{-3mm}
\end{equation*}

\subsection{Analysis Results}
\label{sec:existence}

\newpara{Speaker-utterance attribution collapses to chance in open-source AVLLMs.}
As shown in \cref{tab:cardinality}, all three open-source AVLLMs perform at or near chance level across all speaker counts, indicating that they largely fail to recover the true speaker--utterance correspondence. In contrast, Gemini 3.7 Flash achieves substantially higher accuracy.

\newpara{Failures are systematic, not random.}
The ordinal-matching rate of the open-source AVLLMs is substantially above chance for all $N\in\{4,5,6\}$, with stronger agreement under row-major ($\mathrm{OM}_{\sigma_\mathrm{row}}$) than column-major order ($\mathrm{OM}_{\sigma_\mathrm{col}}$). That is, rather than relying on audio-visual cues, the models associate utterances with speakers by aligning the temporal order of utterances with the spatial order of speakers in row-major layout, which we attribute to the ordinal-matching bias.

\subsection{Causal Validation via Positional Intervention}
\label{sec:causal}
The preceding results provide observational evidence of ordinal-matching bias. To establish causality, we manipulate the model's visual position encodings of each speakers and test whether this intervention alters model's predictions. 

\newpara{Positional interventions.}
If the model relies on genuine audio--visual correspondence, altering the visual position encodings of speakers should have little effect. If it instead relies on ordinal-matching bias, its predictions should shift accordingly, aligning the temporal order of utterances with the intervened visual ordering: the ordinal-matching rate measured against the original on-screen ordering ($\sigma$) should drop, while the rate measured against the intervened position ordering ($\sigma^\dagger$) remains high.
Concretely, each video patch enters the vision encoder with an integer
coordinate $(h,w)$ within its frame, from which its positional encoding is
derived. We rewrite these coordinates per speaker region, identically across frames, in two ways. \emph{Swap} exchanges the positional assignments of the top-left and top-right speaker regions, whereas \emph{Rotate} cyclically permutes the assignments of all four regions, such that each region receives the coordinates of its clockwise neighbor. For each intervention, we measure ordinal-matching rate with respect to two ordinal templates: the original row-major ordering of speakers on the screen ($OM_{\sigma_{row}}$) and the ordering implied by the intervened positions ($OM_{\sigma_{row}^\dagger}$).

\definecolor{gainblue}{RGB}{40,100,190}
\definecolor{lossred}{RGB}{190,60,60}
\newcommand{\dnn}[1]{#1\rlap{\kern1.5pt\textcolor{lossred}{$\scriptstyle\downarrow$}}}
\newcommand{\upp}[1]{\textbf{#1}\rlap{\kern1.5pt\textcolor{gainblue}{$\scriptstyle\uparrow$}}}
\newcommand{\OMrow}{$\mathrm{OM}_{\sigma_{\mathrm{row}}}$}
\newcommand{\OMrowd}{$\mathrm{OM}_{\sigma_{\mathrm{row}}^{\dagger}}$}

\begin{table}[t]
\centering
\footnotesize
\setlength{\tabcolsep}{2pt}
\renewcommand{\arraystretch}{1.05}
\caption{\textbf{Causal evidence of ordinal-matching bias} ($N{=}4$). We intervene on the visual positional indices while preserving the original on-screen order (Swap, Rotate). Relative to the baseline, $\mathrm{OM}_{\sigma_{\mathrm{row}}}$, defined with respect to the original on-screen order $\sigma_{\mathrm{row}}$, drops, whereas $\mathrm{OM}_{\sigma_{\mathrm{row}}^{\dagger}}$, defined with respect to the injected coordinate order $\sigma_{\mathrm{row}}^{\dagger}$, stays high.}
\vspace{-3mm}
\begin{tabular}{@{}ll
>{\centering\arraybackslash}m{0.8cm}
>{\centering\arraybackslash}m{0.8cm}
>{\centering\arraybackslash}m{0.8cm}
>{\centering\arraybackslash}m{0.8cm}
>{\centering\arraybackslash}m{0.8cm}
>{\centering\arraybackslash}m{0.8cm}
@{\hspace{3pt}}}
\toprule
& & \multicolumn{2}{c}{\makebox[0pt][c]{\scriptsize Qwen2.5-Omni\kern3mm}}
& \multicolumn{2}{c}{\makebox[0pt][c]{\kern1mm\scriptsize video-SALMONN2+}}
& \multicolumn{2}{c}{\makebox[0pt][c]{\kern3mm\scriptsize Qwen3-Omni}} \\
\cmidrule(lr){3-4} \cmidrule(lr){5-6} \cmidrule(l){7-8}
\textbf{Interv.} & \textbf{Metric} & SW & WS & SW & WS & SW & WS \\
\midrule
Baseline & \OMrow
& 70.4 & 75.0 & 67.4 & 53.0 & 68.4 & 59.4 \\
\midrule
\multirow{2}{*}{Swap}
& \OMrow
& \dnn{28.0} & \dnn{32.9} & \dnn{39.5} & \dnn{36.4} & \dnn{34.6} & \dnn{32.4} \\
& \OMrowd
& \upp{68.8} & \upp{75.5} & \upp{68.9} & \upp{55.9} & \upp{61.5} & \upp{58.2} \\
\midrule
\multirow{2}{*}{Rotate}
& \OMrow
& \dnn{24.8} & \dnn{15.0} & \dnn{28.0} & \dnn{19.8} & \dnn{29.9} & \dnn{28.1} \\
& \OMrowd
& \upp{56.2} & \upp{61.4} & \upp{55.9} & \upp{57.9} & \upp{43.1} & \upp{38.0} \\
\bottomrule
\end{tabular}
\label{tab:intervention}
\vspace{-6mm}
\end{table}

\begin{table*}[t]
\centering
\footnotesize
\setlength{\tabcolsep}{4pt}
\renewcommand{\arraystretch}{1.05}
\caption{\textbf{OD-FT suppresses ordinal-matching bias and improves real-world understanding.} OD-FT improves accuracy on the synthetic diagnostic corpus while reducing ordinal-matching bias to near-chance levels, and consistently improves performance across all three real-world benchmarks. In contrast, OC-FT provides little improvement on the synthetic corpus, exacerbates ordinal-matching bias, and produces smaller gains on real-world benchmarks.}
\vspace{-2mm}
\label{tab:ft}
\resizebox{0.95\textwidth}{!}{%
\begin{tabular}{@{}l
cccc!{\vrule width 0.2pt} cccc!{\vrule width 0.2pt} cccc
!{\vrule width 1pt} ccc@{}}
\toprule
& \multicolumn{12}{c!{\vrule width 1pt}}{\textbf{Synthetic diagnosis corpus}}
& \multicolumn{3}{c}{\textbf{Real-world benchmarks}} \\
\cmidrule(lr){2-13} \cmidrule(l){14-16}
& \multicolumn{4}{c}{$N=4$} & \multicolumn{4}{c}{$N=5$} & \multicolumn{4}{c!{\vrule width 1pt}}{$N=6$}
& \makecell{Social\\Omni} & \makecell{Daily\\Omni} & \makecell{AV\\Speaker} \\
\cmidrule(lr){2-5} \cmidrule(lr){6-9} \cmidrule(lr){10-13}
\cmidrule(lr){14-14}\cmidrule(lr){15-15}\cmidrule(l){16-16}
& \multicolumn{2}{c}{Accuracy $\uparrow$} & \multicolumn{2}{c}{$\mathrm{OM}_{\sigma_{row}}$ $\downarrow$}
& \multicolumn{2}{c}{Accuracy $\uparrow$} & \multicolumn{2}{c}{$\mathrm{OM}_{\sigma_{row}}$ $\downarrow$}
& \multicolumn{2}{c}{Accuracy $\uparrow$} & \multicolumn{2}{c!{\vrule width 1pt}}{$\mathrm{OM}_{\sigma_{row}}$ $\downarrow$}
& Acc. $\uparrow$ & Acc. $\uparrow$ & Acc. $\uparrow$ \\
\cmidrule(lr){2-3}\cmidrule(lr){4-5}\cmidrule(lr){6-7}\cmidrule(lr){8-9}\cmidrule(lr){10-11}\cmidrule(lr){12-13}
\textbf{Model}
& SW & WS & SW & WS
& SW & WS & SW & WS
& SW & WS & SW & WS
& & & \\
\midrule
Qwen2.5-Omni
& \underline{26.9} & \underline{25.9} & \underline{70.4} & \underline{75.0}
& \underline{23.7} & \underline{24.1} & \underline{54.6} & \underline{62.2}
& \underline{18.9} & \underline{19.3} & \underline{48.9} & \underline{53.7}
& 38.9 & 64.3 & 44.8 \\
\quad + OC-FT
& 25.0 & 25.8 & 98.5 & 98.1
& 20.3 & 21.7 & 78.3 & 69.6
& 17.2 & 18.1 & 66.6 & 58.7
& \underline{39.5} & \underline{67.2} & \underline{45.8} \\
\rowcolor{modelhighlight}
\quad + \textbf{OD-FT}
& \textbf{97.0} & \textbf{97.9} & \textbf{26.5} & \textbf{25.9}
& \textbf{94.8} & \textbf{92.4} & \textbf{21.5} & \textbf{20.3}
& \textbf{91.9} & \textbf{83.7} & \textbf{19.1} & \textbf{16.4}
& \textbf{54.8} & \textbf{69.6} & \textbf{48.4} \\
\midrule
video-SALMONN2+
& \underline{34.0} & 27.8 & \underline{67.4} & \underline{53.0}
& 27.7 & 25.0 & \underline{60.6} & \underline{49.4}
& 27.0 & 24.7 & \underline{49.5} & \underline{46.3}
& 43.2 & 65.1 & 44.7 \\
\quad + OC-FT
& 34.0 & \underline{31.9} & 80.6 & 67.0
& \underline{32.4} & \underline{31.9} & 63.7 & 60.2
& \underline{39.9} & \underline{39.4} & 55.1 & 51.2
& \underline{44.6} & \underline{67.8} & \textbf{47.0} \\
\rowcolor{modelhighlight}
\quad + \textbf{OD-FT}
& \textbf{75.0} & \textbf{45.9} & \textbf{29.8} & \textbf{24.6}
& \textbf{79.2} & \textbf{53.5} & \textbf{22.9} & \textbf{21.9}
& \textbf{89.9} & \textbf{80.8} & \textbf{18.5} & \textbf{15.9}
& \textbf{45.6} & \textbf{68.2} & \underline{46.9} \\

\bottomrule
\vspace{-8mm}
\end{tabular}
}
\end{table*}
\newpara{Intervention results.}
As shown in \cref{tab:intervention}, across all intervention types and models, $OM_{\sigma_{\mathrm{row}}}$ drops, while $OM_{\sigma_{\mathrm{row}}^\dagger}$ have high value; that is, the models' predictions follow the rewritten positions rather than the actual on-screen layout. This provides causal evidence for ordinal-matching bias.

\section{Mitigating Ordinal-Matching Bias}
\label{sec:balanced_ft}

Having established that speaker-utterance attribution failures of AVLLMs stem from ordinal-matching bias, we investigate whether this bias can be mitigated through counterfactual fine-tuning (\cref{sec:balanced_method}) and whether this improves real-world benchmark performance (\cref{sec:exp}).

\subsection{Ordinal-Decoupled Fine-tuning Experiments}
\label{sec:balanced_method}

\newpara{Fine-tuning corpus.} We construct a synthetic fine-tuning corpus of 400 videos spanning two- and four-speaker layouts. Unlike the diagnostic dataset in \cref{sec:diag_setup}, these videos feature human speakers. In \emph{Ordinal-Decoupled Fine-Tuning} (OD-FT), we randomly permute the utterance order independently of the speakers' spatial order, so that a speaker's visual position does not correspond to their utterance order. To isolate the effect of this decoupling from fine-tuning itself, we additionally construct a control corpus, \emph{Ordinal-Coupled Fine-Tuning} (OC-FT), using the same visual and audio content but preserving strict ordinal correspondence: the $k$-th speaker in row-major order always produces the $k$-th utterance.

\newpara{Implementation details.}
We fine-tune Qwen2.5-Omni (7B) and video-SALMONN2+ (7B) using LoRA~\cite{hu2022lora} with rank 8 and $\alpha{=}16$. We optimize the models with AdamW using a learning rate of $10^{-4}$. Training is performed with a batch size of 1 and 8 gradient accumulation steps for two epochs.

\subsection{Experimental Results}
\label{sec:exp}
\newpara{OD-FT suppresses ordinal-matching bias.} We repeat the evaluation from \cref{sec:existence} to test whether ordinal-matching bias is reduced after fine-tuning. As shown in the synthetic-corpus results of \cref{tab:ft}, OD-FT improves accuracy while reducing the ordinal-matching rate to near-chance levels. In contrast, OC-FT yields little improvement in accuracy and further strengthens the ordinal-matching bias.

\label{sec:balanced_diagnostic}

\label{sec:realworld}

\newpara{OD-FT improves general performance on real-world datasets.} We further test whether reducing ordinal-matching bias improves reasoning performance on real-world datasets. We evaluate on three conversation-centric audio--visual benchmarks: AVSpeaker~\cite{nguyen2025see}, DailyOmni~\cite{zhou2025daily}, and the speaker-perception subset of SocialOmni~\cite{xie2026socialomni}. As shown in the real-world benchmark results of  \cref{tab:ft}, OD-FT improves performance across the three benchmarks, yielding average gains of 8.27\% for Qwen2.5-Omni and 2.57\% for video-SALMONN2+. It also generally outperforms OC-FT, suggesting that these gains are not merely a generic effect of fine-tuning but specifically arise from breaking the correspondence between speaker position and utterance order. 

\begin{figure}[t]
\centering
\includegraphics[width=0.95\linewidth]{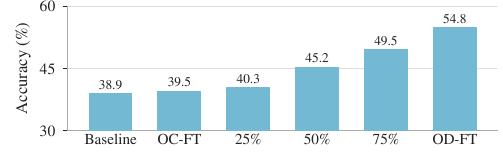}
\vspace{-3mm}
\caption{\textbf{Effect of the decoupled share.} SocialOmni accuracy increases with the fraction of decoupled examples in the fine-tuning corpus, from 0\% (OC-FT) to 100\% (OD-FT).}
\vspace{-3mm}
\label{fig:dose}
\end{figure}

\newpara{Ablation on the decoupled share.}
We fine-tune Qwen2.5-Omni on mixtures of the OC-FT and OD-FT corpora, varying the proportion of decoupled examples (25\%, 50\%, and 75\%). As shown in \cref{fig:dose}, SocialOmni accuracy increases with the decoupled share, further confirming that the gain comes from breaking the ordinal correspondence.

\section{Conclusion}
\label{sec:conclusion}
Current open-source AVLLMs remain unreliable at attributing utterances to
the correct visible speaker. We show that this failure is not random:
models follow a systematic \emph{ordinal-matching bias}, pairing the
$k$-th utterance with the $k$-th speaker in visual position order.
Importantly, this bias can be mitigated with minimal supervision.
\emph{Ordinal-Decoupled Fine-Tuning} on only 400 synthetic videos
suppresses the bias, and the resulting improvements transfer to three
real-world benchmarks. Although our evaluation centers on structured environments and clean speech, we leave dynamic multi-party extensions and root-cause analyses to future work. Ultimately, this study highlights the need for AVLLMs built on genuine audio--visual alignment rather than fragile shortcuts.
\bibliographystyle{IEEEbib}
\bibliography{shorstrings,strings}

\end{document}